\documentclass[a4paper, 12pt, twoside]{article}
\input{jadtstyle.cls}

\usepackage{graphicx}
\usepackage{pgfplots}
\usetikzlibrary{shapes.geometric, arrows.meta, positioning, fit, backgrounds}
\pgfplotsset{compat=1.18}
\usepackage{amsmath}
\usepackage{amssymb}
\usepackage{booktabs}
\usepackage{url}

\usepackage{xcolor}
\usepackage[normalem]{ulem}

\title{\fontsize{18}{21.6}\textbf{Beyond frequency measures: Can contextual embeddings capture meaning change in scientific texts?}}

\author[1]{Jianying Liu}
\author[2]{Kim Gerdes}
\author[3]{Jean-Marc Deltorn}

\affil[1]{LISN, Université Paris-Saclay; CEIPI, Université de Strasbourg $-$ jianying.liu@universite-paris-saclay.fr}
\affil[2]{LISN, Université Paris-Saclay $-$ kim.gerdes@universite-paris-saclay.fr}
\affil[3]{CEIPI, Université de Strasbourg $-$ jm.deltorn@ceipi.edu}

\date{\vspace*{-3em}}

\begin{document}

\maketitle

\fancyhead[CE]{\textsc{\footnotesize{J. Liu, K. Gerdes, J.M. Deltorn}}}
\fancyhead[CO]{\textsc{\footnotesize{Beyond frequency measures: Can contextual embeddings capture meaning change in scientific texts?}}}

\selectlanguage{english} 

\setlength{\parindent}{\z@}
\setlength{\baselineskip}{11pt}


\vspace*{2pt}
\abstract*{Abstract}

\begin{footnotesize}{
Identifying technological trends is a core scientometric task, yet traditional frequency-based approaches struggle to capture substantial meaning shifts of domain-specific terms. We hypothesise that contextual embeddings can complement frequency dynamics to effectively track diachronic semantic change. We compare frequency and embedding-based approaches across Astrophysics and NLP corpora spanning from 2010 to 2024. Candidate terms are extracted using KeyBERT (utilizing SciBERT as its underlying language model) and filtered for significant frequency increases using Fisher’s exact test. These terms are then evaluated for genuine semantic shift by domain experts to establish ground-truth labels. To quantify semantic drift, each term's contextual embedding ``clouds'' from the two discrete periods are compared using multiple metrics: cosine distance, average pairwise distance, Hotelling-type $T^2$, and maximum mean discrepancy. Results indicate that frequency-based methods align slightly better with human judgments of ``trend-related terms'' than semantic metrics (Precision@50 of 0.62 vs 0.60 in Astrophysics). The two signals show a correlation of around 0.6. Several terms identified exclusively by embedding metrics (e.g., ``primordial black holes'') represent critical conceptual developments invisible to pure frequency analysis. These findings indicate that semantic metrics may capture complementary information, highlighting the value of integrating contextual embeddings into scientometric trend analysis.}\end{footnotesize}

\vspace*{14pt}{\textbf{Keywords:}} \begin{footnotesize}{contextual embeddings, Lexical semantic change, Large/pretrained language model, Embedding distance metrics, Diachronic statistical analysis of scientific terms}\end{footnotesize}


\setlength{\parskip}{6pt}
\setlength{\baselineskip}{14pt}

\section{Introduction}

Identifying technological hot spots and trends from large textual corpora is a core task in scientometrics. Traditional analysis mainly relies on frequency-based indicators and co-word networks \citep{callonCowordAnalysisTool1991}, which provide useful macroscopic views but remain limited in capturing substantial meaning shifts of domain-specific key terms over time.

In this work, we hypothesize that diachronic semantic change of in-domain key terms can occur independently of term frequency dynamics and can be captured by contextual vector representations, thereby providing complementary information. To test this hypothesis, we conduct a comparative study between frequency-based and contextual embedding-based approaches in two scientific domains: Astrophysics and speech and natural language processing (SNLP), using article corpora from 2010 and 2024 for Astrophysics, and from 2010 and 2020 for SNLP.

This work's contributions are fourfold: (1) we propose a trend-detection method using contextual embeddings to avoid full-corpus LLM processing, and we show how its performance varies across domains; (2) we evaluate keyword quality and the difficulty of semantic-shift annotation via human and LLM annotations; (3) we compare various embedding distance metrics against frequency-based methods and analyze the correlation between these two types of metrics; and (4) we qualitatively analyze categories of semantic shifts in scientific domains, with examples from astrophysics and SNLP.

\section{Related Work}

\subsection{Diachronic Embedding Models}
Detecting semantic change using vector representations has evolved from static to contextual frameworks. Early works predominantly utilized static dense word vectors trained independently and aligned post-hoc \citep{hamiltonDiachronicWordEmbeddings2016}, or enforced temporal alignment during representation learning \citep{dicarloTrainingTemporalWord2019, bianchiCompassalignedDistributionalEmbeddings2020}. Recently, however, contextualized models like BERT and RoBERTa have dominated lexical semantic change (LSC) tasks \citep{schlechtwegSemEval2020Task2020}. For scientific literature, domain-adapted language models such as SciBERT \citep{beltagySciBERTPretrainedLanguage2019} demonstrate superior semantic tracking capabilities.

\subsection{Semantic Change Assessment Metrics}
To quantify semantic drift, LSC methodologies typically operate under two paradigms: prototype tracking (e.g., Cosine Distance) and distribution tracking \citep{peritiLexicalSemanticChange2024}. The latter models the full sense inventory by clustering contextual embeddings across time (e.g., APD). Beyond these, metrics like the Regularized Hotelling statistic $T^2$ stabilized by Ledoit--Wolf shrinkage \citep{ledoit2004well} and Maximum Mean Discrepancy (MMD) \citep{grettonKernelTwoSampleTest2012, jayasumana2024rethinking} have gained traction for high-dimensional Mahalanobis and distribution distance estimation \citep{hanConditionalWordEmbedding2018, podolskiy2021revisiting}.

Beyond these, some studies have also used the Regularized Hotelling statistic $T^2$ for distribution distance. While relatively niche in NLP \citep{hanConditionalWordEmbedding2018}, it is a standard approach in gene set analysis for high-dimensional data \citep{chen2010two}, especially when stabilized by Ledoit--Wolf shrinkage \citep{robinson2022improvement, ledoit2004well}. Fundamentally, $T^2$ represents the Mahalanobis distance between mean vectors, a metric extensively applied in NLP tasks like Out-Of-Distribution (OOD) detection \citep{podolskiy2021revisiting}. Furthermore, Maximum Mean Discrepancy (MMD) \citep{grettonKernelTwoSampleTest2012} measures the distance between two embedding distributions and can also be used to align latent spaces. It has been successfully applied in both computer vision \citep{jayasumana2024rethinking, baktashmotlagh2016distribution} and NLP \citep{fonseca2020learning}.

The choice of distance metrics is critical, as relying solely on cosine distance can be insufficient for robust semantic shift detection. \citet{azarpanah2021measuring} demonstrate that the selection of similarity measures and descriptive statistics (e.g., min, max, mean, median) significantly influences the conclusions of word embedding association tests. To address these sensitivities, \citet{liu-etal-2021-statistically} incorporate non-parametric permutation tests with contextual embeddings, ensuring that observed semantic shifts are statistically significant and not artifacts of sampling variance.

\subsection{Correlation Between Frequency and Semantic Shift}
The interaction between frequency trajectories and meaning change remains a debated null hypothesis. \citet{hamiltonDiachronicWordEmbeddings2016} proposed statistical laws of semantic change (conformity and innovation), which were later challenged as potential artifacts of static embeddings and simple cosine distances \citep{dubossarskyOuttaControlLaws2017}. Subsequent work \citep{keidarSlangvolutionCausalAnalysis2022} using causal DAG modeling observed a decoupling between volatile frequency changes and genuine conceptual drift, justifying the exploration of advanced quantitative techniques to isolate substantive semantic evolution in scientific corpora.

\section{Methodology}

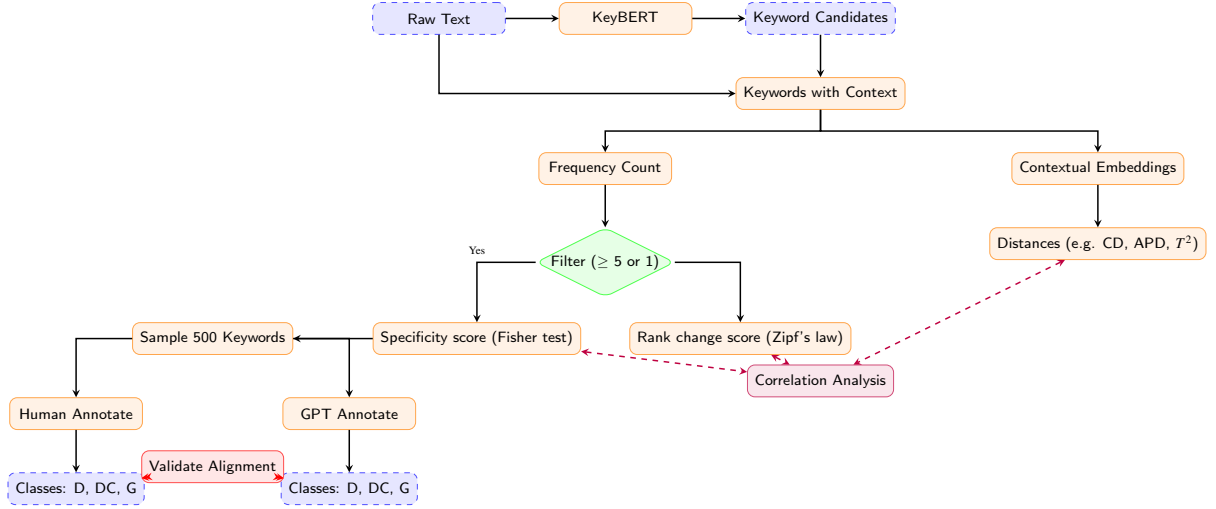
\begin{figure}[h!]
    \centering
    \resizebox{\textwidth}{!}{
    \begin{tikzpicture}[
        node distance=0.8cm and 1.0cm,
        base/.style={rectangle, rounded corners, draw=black, minimum width=2.5cm, minimum height=0.6cm, text centered, font=\sffamily\scriptsize},
        io/.style={base, fill=blue!10, draw=blue!70, dashed},
        process/.style={base, fill=orange!10, draw=orange!70},
        decision/.style={base, diamond, aspect=2, fill=green!10, draw=green!70, inner sep=0pt},
        arrow/.style={thick, ->, >=stealth}
    ]

    \node (text) [io] {Raw Text};
    \node (keybert) [process, right=of text] {KeyBERT};
    \node (candidates) [io, right=of keybert] {Keyword Candidates};
    \node (context) [process, below=of candidates] {Keywords with Context};

    \node (freq) [process, below left=of context, xshift=-0.2cm] {Frequency Count};
    \node (filter) [decision, below=of freq] {Filter ($\ge$ 5 or 1)};
    
    \node (fisher) [process, below left=of filter, xshift=1.2cm] {Specificity score (Fisher test)};
    \node (zipf) [process, below right=of filter, xshift=-1.2cm] {Rank change score (Zipf's law)};
    \node (emb) [process, below right=of context, xshift=1cm] {Contextual Embeddings};
    \node (dist) [process, below=of emb] {Distances (e.g. CD, APD, $T^2$)};

    \node (sample) [process, left=of fisher, xshift=-0.5cm] {Sample 500 Keywords};
    \node (human) [process, below left=of sample, xshift=1.2cm] {Human Annotate};
    \node (gpt) [process, below right=of sample, xshift=-1.2cm] {GPT Annotate};
    \node (human_classes) [io, below=of human] {Classes: D, DC, G};
    \node (gpt_classes) [io, below=of gpt] {Classes: D, DC, G};
    
    \node (validate) [process, fill=red!10, draw=red!70, below=of sample, yshift=-1.0cm] {Validate Alignment};
    
    \node (corr) [process, fill=purple!10, draw=purple!70, below=of context, yshift=-4.0cm] {Correlation Analysis};

    \draw [arrow] (text) -- (keybert);
    \draw [arrow] (keybert) -- (candidates);
    \draw [arrow] (candidates) -- (context);
    \draw [arrow] (text.south) |- (context.west);

    \draw [arrow] (context.south) -- ++(0,-0.4) -| (freq.north);
    \draw [arrow] (context.south) -- ++(0,-0.4) -| (emb.north);

    \draw [arrow] (freq) -- (filter);
    \draw [arrow] (filter) -| (fisher) node[pos=0.5, above, font=\tiny] {Yes};
    \draw [arrow] (filter) -| (zipf);
    
    \draw [arrow] (emb) -- (dist);

    \draw [arrow] (fisher) -- (sample);
    \draw [arrow] (sample) -| (human);
    \draw [arrow] (sample) -| (gpt);
    \draw [arrow] (human) -- (human_classes);
    \draw [arrow] (gpt) -- (gpt_classes);

    \draw [arrow, <->, dashed, red] (human_classes) -- (validate);
    \draw [arrow, <->, dashed, red] (gpt_classes) -- (validate);

    \draw [arrow, <->, dashed, purple] (fisher) -- (corr);
    \draw [arrow, <->, dashed, purple] (zipf) -- (corr);
    \draw [arrow, <->, dashed, purple] (dist) -- (corr);

    \end{tikzpicture}
    }
    \caption{Processing pipeline and methodology flowchart including semantic extraction, dual-metric distance comparison, and LLM annotation validation steps.}
    \label{fig:methodology_diagram}
\end{figure}

\subsection{Data Sources}
We test our hypothesis on two distant scientific domains: Astrophysics (Astro) and Speech and Natural Language Processing (SNLP). For both corpora, only article titles and abstracts are utilized. We sample data from two distinct time periods for comparative analysis (2010 vs. 2024 for Astrophysics, and 2010 vs. 2020 for SNLP). \footnote{The time window asymmetry between the two corpora does not invalidate the comparison: Astrophysics appears to evolve more slowly and needs a longer interval for semantic shifts to emerge, whereas SNLP changed more rapidly during the 2010s, so substantial shifts are already visible within 2010--2020. We therefore treat these windows as domain-specific observation periods.} The Astrophysics corpus is built from arXiv articles labeled \texttt{astro-ph}, yielding 32,336 articles. The SNLP domain reuses the NLP4NLP corpus from \citep{marianiNLP4NLP+5DeepRevolution2022}, scoped to 8,413 English papers. Table \ref{tab:data} summarizes the corpus statistics.

\begin{table}[h]
\centering
\caption{Corpus statistics: Astrophysics (Astro) and SNLP}
\resizebox{\textwidth}{!}{
\begin{tabular}{llrrr}
\toprule
\textbf{Domain} & \textbf{Statistic} & \textbf{2010} & \textbf{2020 / 2024} & \textbf{Total} \\
\midrule
Astrophysics & Paper number & 13,118 & 19,218 & 32,336 \\
 & Token count & 2,565,045 & 4,186,388 & 6,751,433 \\
 & Keyword candidates (forms) & 64,017 & 97,050 & 147,728 (union) \\
 & Studied keyterms ($occ \ge 5$) & \multicolumn{3}{r}{8,013} \\
\midrule
SNLP & Paper number & 3,274 & 5,139 & 8,413 \\
 & Token count & 445,562 & 824,357 & 1,269,919 \\
 & Keyword candidates (forms) & 19,706 & 29,537 & 46,572 (union) \\
 & Studied key terms ($occ \ge 1$) & \multicolumn{3}{r}{6,140} \\
\bottomrule
\end{tabular}
}
\label{tab:data}
\end{table}

\subsection{Keyword Extraction and Embedding}
We perform a two-step domain-specific keyword extraction (Figure \ref{fig:methodology_diagram}). First, KeyBERT \citep{grootendorst2021maartengr}, configured with SciBERT, extracts the top 10 n-grams per article to act as candidate terms. Instead of restricting our tracking strictly to articles where a term was flagged as important, we consolidate the unique candidates and perform a global parse to extract their occurrence counts ($occ$) across all texts in the two distinct years. Filtering out extremely rare terms with an occurrence threshold yields an effective vocabulary of 8,013 terms for Astrophysics and 6,140 for SNLP. We then apply SciBERT to compute contextual embeddings for each valid term, using a surrounding context window of 125 words ($ws=125$) on each side.

Ultimately, between the initial time period ($t_1$) and the target time period ($t_2$), each term $X$ is represented by its total occurrences ($occ_1$ and $occ_2$), its popularity ranks sorted descendingly by occurrences ($rank_1$ and $rank_2$), and two temporal sets of contextual embeddings ($\Phi_1$ and $\Phi_2$) complete with their respective mean prototypes ($\mu_1$ and $\mu_2$).

\subsection{Popularity Metrics}
We measure term popularity change using two frequency-based methods. 
The first relies on Zipf's law to attenuate sensitivities caused by drastic rank scaling compared to most frequent words. To avoid noise in the long tail of the distribution, we fit an inverse power-law coefficient $\alpha$ on the top 500 terms: $\widehat{occ} = 10^b \cdot rank^{-\alpha}$.
A popularity growth ratio, denoted as $\rho$, is then calculated to score the relative usage expansion:
\begin{equation}
\rho = - \left(\frac{rank_2}{rank_1}\right)^\alpha
\end{equation}

Secondly, to rigorously account for the exponential baseline growth of the entire scientific corpus, we measure the specificity and statistical significance of an absolute frequency surge in $t_2$ using Fisher's exact test. For a given term $X$, the probability mass function representing its hypergeometric distribution is formulated as:
\begin{equation}
p(occ_2; N, occ_{\text{all}}, n) = \frac{\binom{occ_{\text{all}}}{occ_2}\binom{N - occ_{\text{all}}}{n-occ_2}}{\binom{N}{n}}
\end{equation}

where $N$ is the total occurrence count of all selected keywords in the corpus, $n$ is the number of selected keywords, and $occ_{\text{all}} = occ_1 + occ_2$ is the total occurrence count of term $X$. The Fisher Specificity Score $F_{spec}$ is derived from the survival function $1-cdf(occ_2)$; higher values indicate statistically significant usage increases over the studied span.

\subsection{Embedding Difference Metrics}
We try both paradigms and different metrics of semantic shift proposed in \citep{peritiLexicalSemanticChange2024}. Under the mean-vector (word-prototype) paradigm, we evaluate both cosine distance and the inverse of cosine similarity (PRT); for the meaning cluster paradigm, we employ average pairwise distance, maximum mean discrepancy, and the Regularized Hotelling statistic $T^2$. Since PRT performs identically to cosine distance, we focus our reporting on cosine distance results.

Let $e_{1,i} \in \Phi_1$ and $e_{2,j} \in \Phi_2$ represent individual contextual embeddings of term $X$ from time periods 1 and 2, with sizes $N_1 = |\Phi_1|$ and $N_2 = |\Phi_2|$ respectively.

\begin{itemize}
    \item \textbf{Cosine Distance:} Directly measured on the mean embeddings $\mu_1$ and $\mu_2$.

    \item \textbf{Average Pairwise Distance (APD):} The average distance between all pairs of embeddings from the two time periods:
    \begin{equation}
    \operatorname{APD}(\Phi_1,\Phi_2) = \frac{1}{N_1 N_2} \sum_{i=1}^{N_1} \sum_{j=1}^{N_2} d(e_{1,i}, e_{2,j})
    \end{equation}

    \item \textbf{Maximum Mean Discrepancy (MMD):} We use a biased MMD estimator with an RBF Gaussian kernel $k(x, y) = \exp(-\gamma \|x-y\|^2)$ and fixed bandwidth $\sigma = 10.0$\footnote{A small ablation over $\sigma \in \{2.5, 5, 10, 20, 40\}$ did not change the main ranking of methods: larger values modestly improved MMD, but it remained clearly weaker than Reg. $T^2$. We therefore keep $\sigma = 10.0$ across the main experiments.}.

    \item \textbf{Regularized Hotelling statistic $T^2$:} We estimate a regularized pooled covariance matrix $\Sigma_{\text{reg}}$ using Ledoit--Wolf shrinkage and measure the separation between sample mean embeddings via:
    \begin{equation}
    T^2 = (\mu_1 - \mu_2)^T \Sigma_{\text{reg}}^{-1} (\mu_1 - \mu_2)
    \end{equation}
\end{itemize}

\subsection{Ground-truth expert annotation}

To establish the human reference labels used in the main evaluation, we curated two samples of 500 candidate keywords, one for Astrophysics and one for SNLP. In each domain, candidates were first ranked by Fisher Specificity Score ($F_{spec}$), then we selected the highest-ranked 200 terms and stratified the remaining 300 across lower-ranked percentiles.

The annotation protocol at this stage used a lightweight guideline, intended to preserve domain-expert judgment on inherently fuzzy boundary cases while remaining simple enough for downstream reliability checks. For each candidate exhibiting salient frequency changes, the annotators judged whether it should be treated as outside the target domain scope (Class G), as an in-domain but semantically stable term (Class D), or as an in-domain term whose contextual or conceptual meaning had changed over time (Class DC). Broad methodological phrases, popular writing formulas, and noisy extractions were generally treated as Class G even when their usage increased, because they do not constitute the target phenomenon of scientific meaning change.

\section{Experiments and Results}

\subsection{Shift Discovery compared to expert judgement}

Because Class DC exclusively represents our target phenomenon of genuine semantic shift, we first benchmark the retrieval performance of each metric specifically against this class. After ranking all candidates by their score according to each metric, we evaluate both the top-$k$ retrieval precision (Precision@$k$) and recall (Recall@$k$), using the human annotations as reference labels.

Table \ref{tab:precision_k} reveals three main patterns.

\textbf{Frequency-based methods align more consistently with human labels overall:} Statistically, frequency-driven methods (popularity growth ratio ($\rho$) and Fisher Specificity Score ($F_{spec}$)) align more consistently with human annotations across the full annotated set. They remain among the top-performing methods across the global ranking, with at least one of them consistently appearing in the top two.

\textbf{Reg. $T^2$ achieves the strongest top-tier precision:} Among all the tested embedding distance paradigms, the Regularized Hotelling statistic (Reg. $T^2$) drastically outperforms alternatives like Cosine distance and MMD. When constrained to high-confidence retrieval windows (e.g., $k \le 50$), the uncalibrated Reg. $T^2$ can even surpass purely frequency-based approaches. It achieves strong top-tier detection (e.g., P@20 of $0.750$ in Astrophysics and $0.700$ in SNLP), aligning closely with human intuition for profound conceptual semantic shift.
However, affected by long-tail effects, this strong early precision does not extend to the full ranking. When all annotated candidates are considered and labelled with a binary Class-DC indicator, only modest Spearman correlation \textbf{$r_s$} is shown between this indicator and the raw Reg. $T^2$, with $r_s=0.132$ for Astro and $r_s=0.129$ for SNLP.
    
\textbf{Naive distances remain relatively insensitive to subtle meaning change:} The smaller retrieval ranges (P@20 and P@50) expose a massive performance gap between the two $T^2$-based metrics and alternative embedding distances, indicating that metric baselines like MMD or raw Cosine mappings are insensitive to the nuanced, high-dimensional meaning shifts recognized by domain experts.

\begin{table}[h]
\centering
\caption{Retrieval performance (Precision/Recall@K) for Class DC and Spearman correlation ($r_s$) against human binary labels. Evaluated across Astrophysics ($N=498$) and SNLP ($N=390$) domains.}
\resizebox{\textwidth}{!}{
\begin{tabular}{lcccccccccc}
\toprule
\multirow{2}{*}{\textbf{Metric}} & \multicolumn{5}{c}{\textbf{Astro (Class DC $N=133$)}} & \multicolumn{5}{c}{\textbf{SNLP (Class DC $N=131$)}} \\
\cmidrule(lr){2-6} \cmidrule(lr){7-11}
 & \textbf{P@20} & \textbf{P@50} & \textbf{P@100} & \textbf{R@100} & \textbf{$r_s$ (Human)} & \textbf{P@20} & \textbf{P@50} & \textbf{P@100} & \textbf{R@100} & \textbf{$r_s$ (Human)} \\
\midrule
Cosine Distance & 0.100 & 0.280 & 0.230 & 0.173 & -0.019 & 0.250 & 0.340 & 0.320 & 0.244 & 0.024 \\
APD (Cosine) & 0.350 & 0.200 & 0.190 & 0.143 & -0.075 & 0.300 & 0.280 & 0.270 & 0.206 & -0.016 \\
MMD & 0.100 & 0.120 & 0.210 & 0.158 & -0.040 & 0.250 & 0.300 & 0.330 & 0.252 & 0.018 \\
Reg. $T^2$ & \textbf{0.750} & \underline{0.600} & 0.400 & 0.301 & 0.132 & \textbf{0.700} & \textbf{0.540} & \textbf{0.460} & \textbf{0.351} & 0.129 \\
Reg. $T^2 \ (1-p\text{-value})$ & 0.350 & 0.340 & 0.320 & 0.241 & 0.168 & 0.400 & 0.420 & 0.420 & 0.321 & \underline{0.129} \\
$\rho$ & \textbf{0.750} & \textbf{0.620} & \textbf{0.550} & \textbf{0.414} & \textbf{0.386} & \underline{0.600} & \underline{0.480} & \underline{0.450} & \underline{0.344} & \textbf{0.176} \\
$F_{spec}$ & \underline{0.500} & 0.520 & \underline{0.460} & \underline{0.346} & \underline{0.344} & 0.550 & 0.400 & \textbf{0.460} & \textbf{0.351} & 0.127 \\
\bottomrule
\end{tabular}
}
\vspace{1mm}
\raggedright {\footnotesize \\ \textit{Note:} Precision and Recall of Class DC annotated by human experts. The total number of Class DC target keywords is 133 for Astrophysics and 131 for SNLP. $r_s$ (Human) represents the Spearman correlation of the metric against the binary indicator for Class DC. Best performing scores are highlighted in \textbf{bold}, and second-best are \underline{underlined}. Same for the following tables.}
\label{tab:precision_k}
\end{table}

To test whether a large Reg. $T^2$ value reflects a robust period difference rather than a sampling artifact, we estimated a permutation-based $p$-value for each keyword by shuffling time period labels 1,000 times, then used $1-p$ as a calibrated score, so that larger values indicate stronger evidence against statistical noise.

Conversely, the statistically calibrated permutation significance $(1 - p\text{-value})$ metric consistently outperforms the raw geometric distance across the full annotated subset ($r_s=0.168$ in Astro, $r_s=0.129$ in SNLP, whereas the corresponding values for the raw geometric distances remain below 0.1 in absolute value). This highlights the ``denoising'' effect of permutation testing: high-dimensional geometric distances are often deceptive, artificially inflated by sampling variance and outlier contexts in small-sample scenarios. Statistical calibration effectively filters this noise, isolating authentic semantic evolution from linguistic background fluctuations. This observation coheres with \citet{liu-etal-2021-statistically}'s assertion that significance testing is important for embedding shift analysis.

\subsection{Correlation Between Metrics}
To understand whether the metrics capture redundant or distinct phenomena, we analyzed the Spearman rank correlation between frequency evolution indicators ($\rho$, $F_{spec}$) and semantic embedding distances within the Class DC subset, i.e., terms confirmed to have undergone semantic meaning change (Table \ref{tab:correlation_dc}).

\begin{table}[h]
\centering
\caption{Cross-metric Spearman Correlation on the \textbf{Class DC Subset}. For the strongest correlation per column (bold), the 95\% confidence interval and statistical significance ($p$-value) are provided in parentheses.}
\resizebox{\textwidth}{!}{
\begin{tabular}{lcccc}
\toprule
\multirow{2}{*}{\textbf{Semantic Metric}} & \multicolumn{2}{c}{\textbf{Astro (Class DC $N=133$)}} & \multicolumn{2}{c}{\textbf{SNLP (Class DC $N=131$)}} \\
\cmidrule(lr){2-3} \cmidrule(lr){4-5}
& \textbf{$\rho$} & \textbf{$F_{spec}$} & \textbf{$\rho$} & \textbf{$F_{spec}$} \\
\midrule
Cosine Distance & 0.161 & -0.290 & 0.135 & -0.264 \\
APD (Cosine) & \underline{0.172} & \underline{0.366} & 0.154 & \underline{0.384} \\
MMD & 0.090 & -0.445 & 0.087 & -0.380 \\
Reg. $T^2$ & \begin{tabular}{@{}c@{}}\textbf{0.646} \\[-0.5ex] \scriptsize ($[0.496, 0.763]$, $p=4.947\times10^{-17}$)\end{tabular} & 0.058 & \begin{tabular}{@{}c@{}}\textbf{0.690} \\[-0.5ex] \scriptsize ($[0.548, 0.789]$, $p=8.314\times10^{-20}$)\end{tabular} & 0.198 \\
Reg. $T^2 \ (1-p\text{-value})$ & 0.155 & \begin{tabular}{@{}c@{}}\textbf{0.489} \\[-0.5ex] \scriptsize ($[0.340, 0.607]$, $p=2.43\times10^{-9}$)\end{tabular} & \underline{0.259} & \begin{tabular}{@{}c@{}}\textbf{0.545} \\[-0.5ex] \scriptsize ($[0.409, 0.658]$, $p=1.723\times10^{-11}$)\end{tabular} \\
\bottomrule
\end{tabular}
}
\vspace{1mm}
\label{tab:correlation_dc}
\end{table}

Our cross-metric analysis reveals a clear divergence in how different methods capture semantic shift. As shown in Table \ref{tab:correlation_dc}, the Reg. $T^2$ statistic demonstrates a strong positive correlation (0.65 for Astro and 0.69 for SNLP) with the absolute frequency growth ($\rho$). In contrast, baseline distance metrics (e.g., Cosine Distance, MMD) show minimal correlation with frequency changes, with APD occasionally being the second highest but still remaining relatively low. This suggests that Reg. $T^2$ acts more as a frequency-sensitive metric: its high values are closely associated with drastic shifts in term frequencies, which geometrically pull the sampling statistics apart.

Conversely, the Reg. $T^2 \ (1 - p\text{-value})$ exhibits a different behavior. Its correlation with absolute frequency growth drops, and it instead aligns more closely with the specificity indicator $F_{spec}$ (0.49 for Astro and 0.55 for SNLP).

These observations suggest two distinct dimensions for tracking terminology evolution. Empirical measurements of shift magnitude (e.g., Reg. $T^2$ and frequency rank drift $\rho$) characterize the physical scale of term displacement, which is often tied to surges in popularity. Meanwhile, statistical measurements (e.g., the Reg. $T^2 (1 - p\text{-value})$ and the $F_{spec}$) act as significance filters. They help control for sampling variance caused by frequency volatility and highlight terms whose newer usage has become more contextually specialized.

\subsection{Annotation Reliability and LLM-Assisted Scalability}

\begin{table}[h]
\centering
\caption{Human--LLM and inter-human annotation consistency.}
\resizebox{\textwidth}{!}{
\begin{tabular}{lcccccccc}
\toprule
\multirow{2}{*}{\textbf{Row}} & \multicolumn{4}{c}{\textbf{Astro}} & \multicolumn{4}{c}{\textbf{SNLP}} \\
\cmidrule(lr){2-5} \cmidrule(lr){6-9}
& \textbf{$N$} & \textbf{Agreement} & \textbf{$\kappa$} & \textbf{DC Agr.} & \textbf{$N$} & \textbf{Agreement} & \textbf{$\kappa$} & \textbf{DC Agr.} \\
\midrule
\texttt{GPT-4.1-mini} & 500 & 0.520 & 0.309 & 0.842 & 390 & 0.610 & 0.409 & 0.634 \\
\texttt{Llama-3.3-70b-versatile} & 496 & 0.643 & 0.449 & 0.654 & 390 & 0.600 & 0.388 & 0.603 \\
\texttt{Qwen3-32b} & 500 & 0.670 & 0.481 & 0.466 & 390 & 0.590 & 0.381 & 0.565 \\
Expert B & -- & -- & -- & -- & 215 & 0.577 & 0.369 & 0.300 \\
Mean inter-LLM & 497 & 0.591 & 0.403 & 0.433 & 390 & 0.658 & 0.481 & 0.664 \\
Mean inter-human & -- & -- & -- & -- & 189 & 0.557 & 0.311 & 0.191 \\
\bottomrule
\end{tabular}
}
\vspace{1mm}
\raggedright {\footnotesize \\ \textit{Note:} The first three rows compare each model against Expert A. The last two rows report the mean pairwise scores across the three listed models and across the three human annotations. \textbf{Agreement} is the proportion of identical labels over the shared keyword set; \textbf{$\kappa$} is Cohen's kappa; \textbf{DC Agr.} is the agreement rate restricted to human-labeled Class DC items.}
\label{tab:annotation_consistency_merged_preview}
\end{table}

Because expert annotation limits the scalability of semantic-shift evaluation, we assess whether LLM-based annotation can support this process with acceptable reliability.

We completed three consistency analyses: Human--LLM comparison for \textit{GPT-4.1-mini}, \textit{Llama-3.3-70b-versatile}, and \textit{Qwen3-32b}; inter-model comparison among the same three models; and inter-annotator comparison. As Table \ref{tab:annotation_consistency_merged_preview} shows, the two domains exhibit different disagreement profiles. Astrophysics shows stronger model-dependent variation, whereas SNLP yields more uniformly moderate agreement across models. Overall, LLM annotation is informative but not stable enough to replace human ground truth.

To diagnose disagreement more precisely, we also conducted a second-round expert agreement study on the subset of annotated SNLP terms occurring more than five times in both periods.

This follow-up uses a refined guideline that decomposes the original decision into three parallel binary judgments: whether a term is in-domain, trend-related, and meaning-shifting. Annotators may additionally mark certainty and assign standardized notes for ambiguous cases, such as truncation, excessive generality, or popular tasks associated with new methods. The study involves the original annotator (Expert A) plus seven additional in-domain experts; Expert A and B completed the full 215-term list, while the others jointly covered a 189-term subset. As Table \ref{tab:annotation_consistency_merged_preview} shows, inter-human agreement is not higher than human-model agreement, confirming that semantic-shift annotation is itself a difficult task.

\subsection{Ablation on Context Window Size}

Our main setting uses $ws=125$, chosen with reference to text length in our corpora: because this study relies on article titles and abstracts only, which together average roughly 200 words, a 125-token window on each side already covers most of the available local context. We also compared results for $ws=5$, $50$, and $125$. The main conclusions remain stable across these settings: frequency-based methods align better with expert labels globally, and Reg. $T^2$ remains the strongest embedding-based retrieval metric. Smaller windows weaken the $T^2$-based scores overall and make APD relatively more correlated with Class DC, suggesting that broader context is beneficial for the statistical distance measures used here.

\section{Discussion: Categories of Semantic Shifts in Scientific Discourse}

\begin{figure}[!ht]
    \centering
    \includegraphics[width=0.24\textwidth]{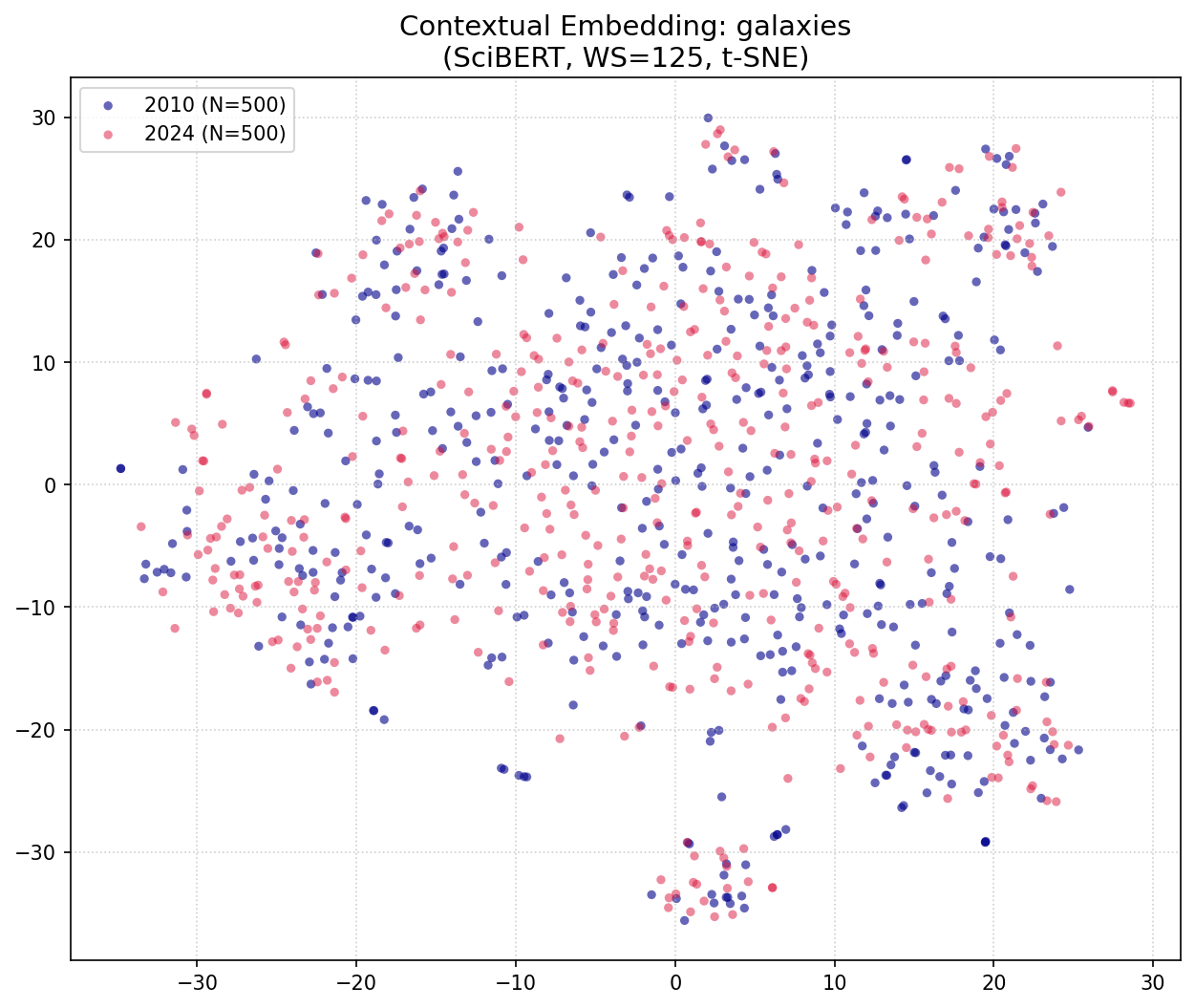}
    \hfill
    \includegraphics[width=0.24\textwidth]{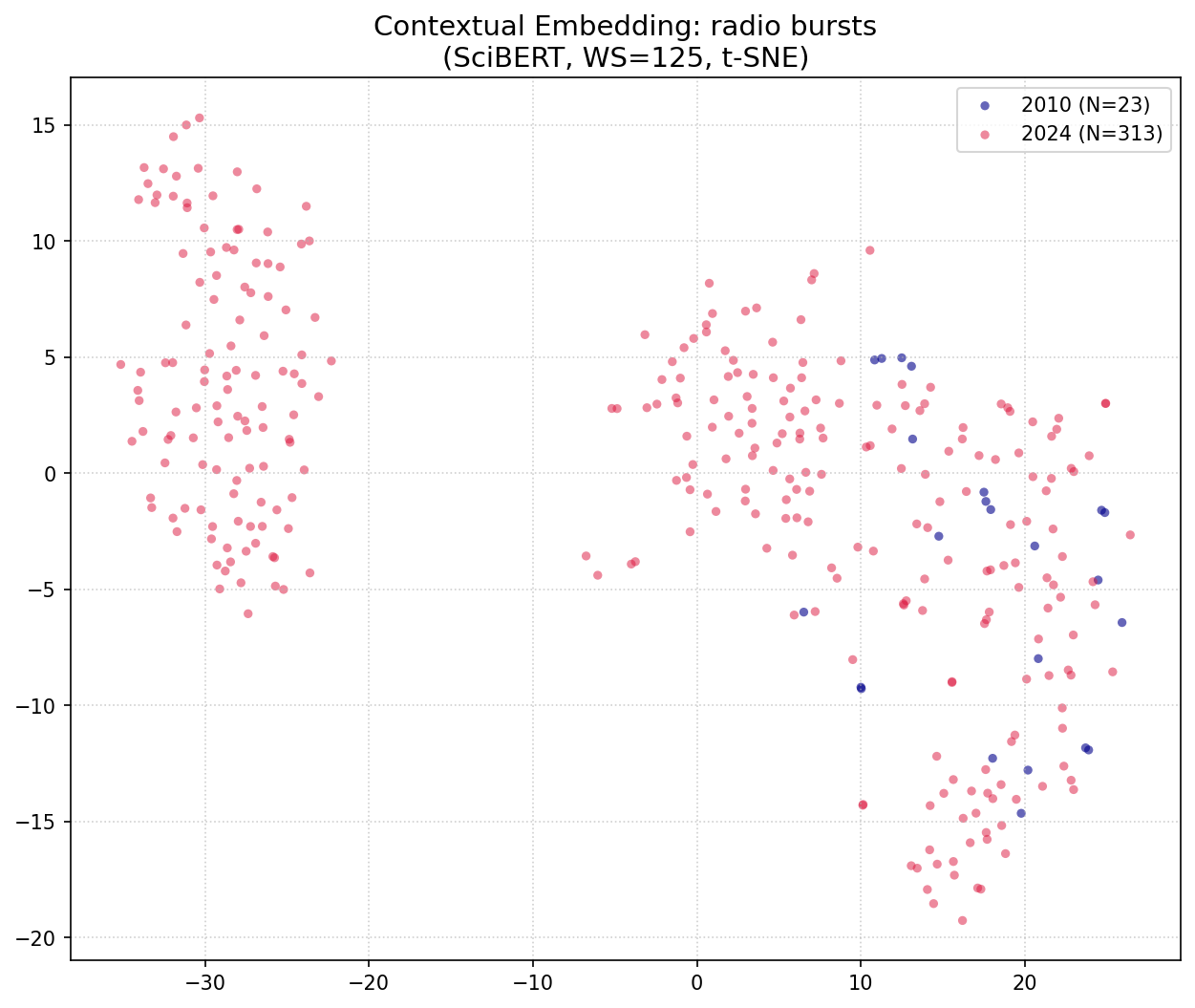}
    \hfill
    \includegraphics[width=0.24\textwidth]{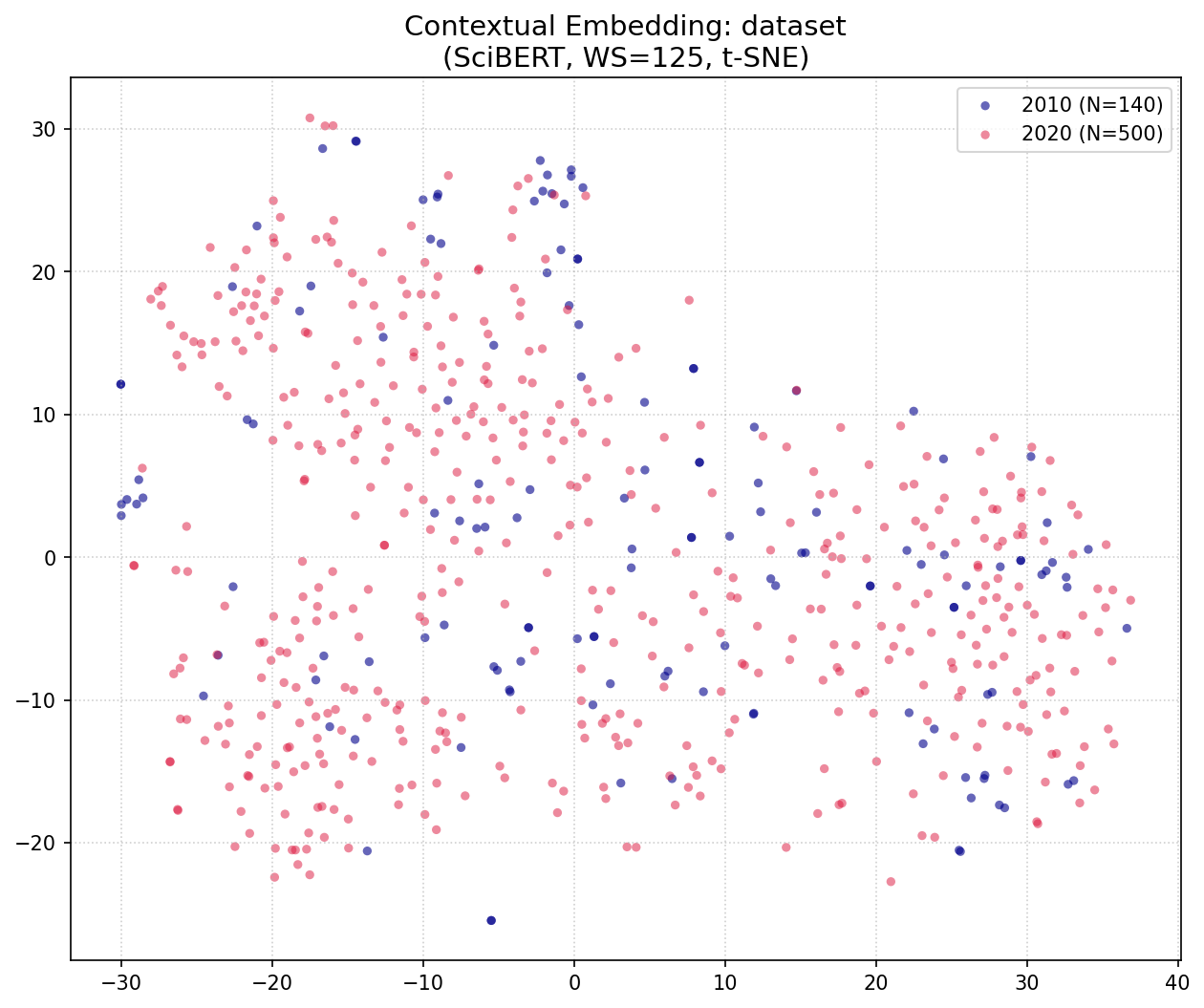}
    \hfill
    \includegraphics[width=0.24\textwidth]{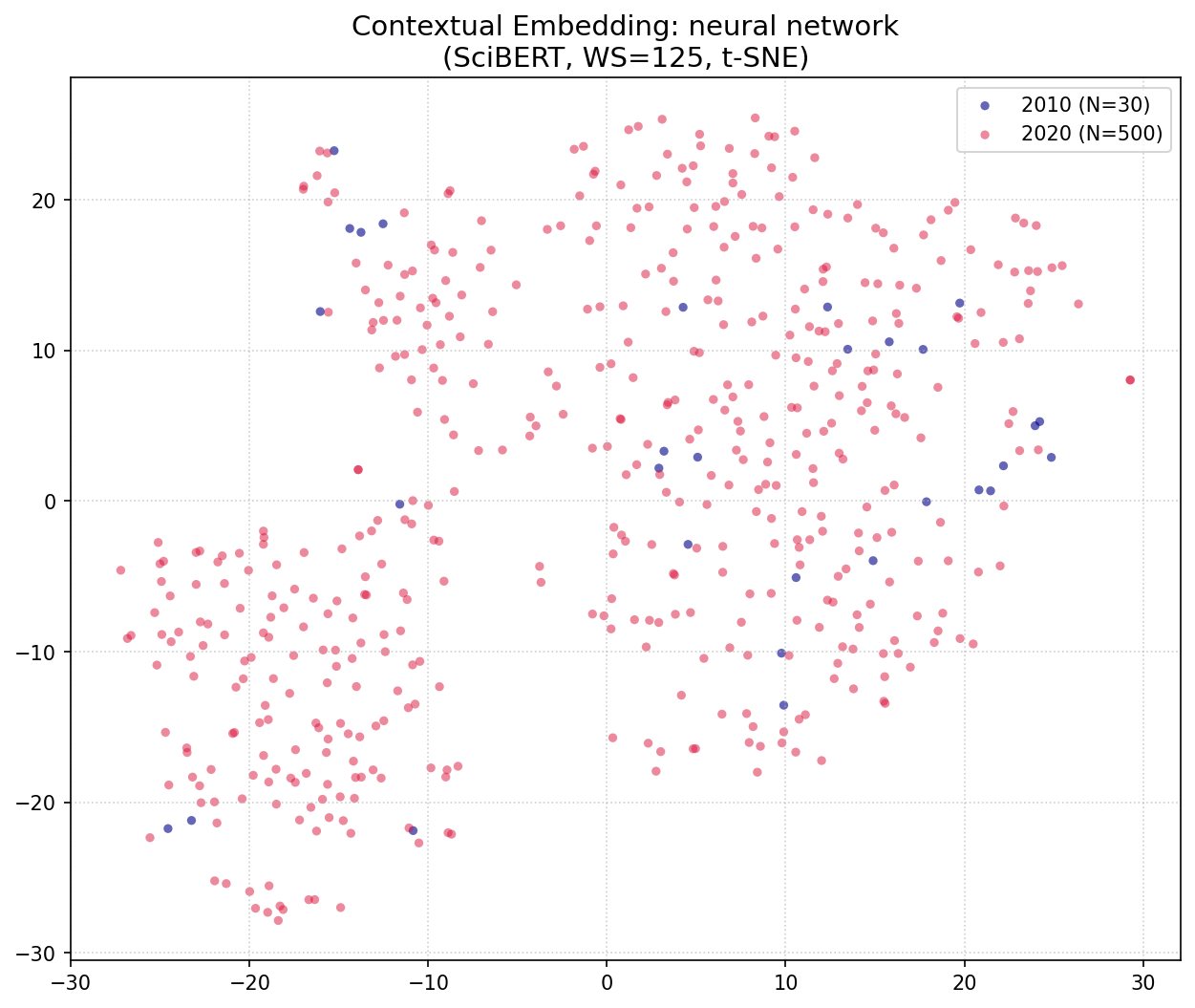}
    \caption{Visualizing the decoupling of frequency growth from semantic shift via contextual embedding tracking. From left to right: \textbf{(A) Astro:} \textit{galaxies} (High Freq, Stable Semantics). \textbf{(B) Astro:} \textit{radio bursts} (High Freq, High Shift). \textbf{(C) SNLP:} \textit{dataset} (High Freq, Stable Semantics). \textbf{(D) SNLP:} \textit{neural network} (High Freq, High Shift). Terms with massive occurrence scale-ups do not automatically incur geometric embedding divergence.}
    \label{fig:scatter_grid}
\end{figure}

A qualitative inspection of high-shift terms demonstrates that semantic evolution in scientific writing follows distinct typological patterns, which are largely decoupled from raw frequency metrics. The scatter plots in Figure \ref{fig:scatter_grid} provide explicit visual evidence that term popularity does not inherently indicate a shift in underlying meaning. For instance, while keywords like \textit{``dataset''} experienced massive frequency growth over the 10-year observational window, their contextual embedding clusters from the two epochs overlap almost entirely, indicating stable phenomenological semantics. Conversely, the contextual coordinates for \textit{``radio bursts''} and \textit{``neural network''} exhibit geometric divergence across the temporal subsets, confirming that their conceptual application has structurally evolved alongside their rise in citation popularity.

Beyond frequency independence, our qualitative review isolates two different modes of semantic change in scientific domains: (1) \textbf{Interpretive Shifts Under a Stable Label}, where the prevailing scientific interpretation fundamentally evolves despite a constant lexical form (e.g., the paradigm shift of \textit{``dark matter''} in Astrophysics, or \textit{``language model''} evolving from statistical $n$-grams to generative architectures in SNLP); and (2) \textbf{Methodological Diffusion}, where a computational tool transitions to widespread foundational deployment, broadening its contextual neighborhood (e.g., the integration of \textit{``machine learning''} in Astrophysics and \textit{``neural network''} in SNLP).

\section{Conclusion and Future Work}
This paper presents a hybrid pipeline combining keyphrase extraction, frequency analytics, and contextual embeddings to detect semantic shifts in scientific corpora. We demonstrate that the correlation between frequency-based indicators and embedding-based scores is highly influenced by the specific metrics chosen. Although frequency statistics provide a strong initial proxy for identifying emerging trends, contextual embeddings uncover unique semantic patterns that occur independently of frequency dynamics. Furthermore, zero-shot LLM annotations show strong domain dependence, making human-in-the-loop validation still indispensable for this task. Future work will extend this framework to finer-grained temporal datasets in order to trace continuous conceptual evolution, distinguish different types of semantic shift more clearly, and compare how they are reflected in contextual embedding distributions.

\bibliography{biblio}         

\begin{thebibliography}{}

\bibitem[Azarpanah and Farhadloo, 2021]{azarpanah2021measuring}
Azarpanah H. and Farhadloo M. (2021).
\newblock Measuring biases of word embeddings: What similarity measures and descriptive statistics to use?
\newblock In {\em Proceedings of the First Workshop on Trustworthy Natural Language Processing}, pp. 8--14.

\bibitem[Baktashmotlagh et~al., 2016]{baktashmotlagh2016distribution}
Baktashmotlagh M., Harandi M., and Salzmann M. (2016).
\newblock Distribution-matching embedding for visual domain adaptation.
\newblock {\em Journal of Machine Learning Research}, 17(108):1--30.

\bibitem[Beltagy et~al., 2019]{beltagySciBERTPretrainedLanguage2019}
Beltagy I., Lo K., and Cohan A. (2019).
\newblock Scibert: A pretrained language model for scientific text.
\newblock In {\em Proceedings of the 2019 conference on empirical methods in natural language processing and the 9th international joint conference on natural language processing (EMNLP-IJCNLP)}, pp. 3615--3620.

\bibitem[Bianchi et~al., 2020]{bianchiCompassalignedDistributionalEmbeddings2020}
Bianchi F., Carlo V.~D., Nicoli P., and Palmonari M. (2020).
\newblock Compass-aligned {Distributional} {Embeddings} for {Studying} {Semantic} {Differences} across {Corpora}.

\bibitem[Callon et~al., 1991]{callonCowordAnalysisTool1991}
Callon M., Courtial J.~P., and Laville F. (1991).
\newblock Co-word analysis as a tool for describing the network of interactions between basic and technological research: {The} case of polymer chemistry.
\newblock {\em Scientometrics}, 22(1):155--205.

\bibitem[Chen and Qin, 2010]{chen2010two}
Chen S.~X. and Qin Y.-L. (2010).
\newblock A two-sample test for high-dimensional data with applications to gene-set testing.
\newblock {\em The Annals of Statistics}, 38(2):808--835.

\bibitem[Di~Carlo et~al., 2019]{dicarloTrainingTemporalWord2019}
Di~Carlo V., Bianchi F., and Palmonari M. (2019).
\newblock Training {Temporal} {Word} {Embeddings} with a {Compass}.
\newblock {\em Proceedings of the AAAI Conference on Artificial Intelligence}, 33(01):6326--6334.

\bibitem[Dubossarsky et~al., 2017]{dubossarskyOuttaControlLaws2017}
Dubossarsky H., Weinshall D., and Grossman E. (2017).
\newblock Outta control: Laws of semantic change and inherent biases in word representation models.
\newblock In {\em Proceedings of the 2017 conference on empirical methods in natural language processing}, pp. 1136--1145.

\bibitem[Fonseca and van Dijk, 2020]{fonseca2020learning}
Fonseca A.~H. and van Dijk D. (2020).
\newblock Learning aligned embeddings for semi-supervised word translation using maximum mean discrepancy.
\newblock {\em arXiv preprint arXiv:2006.11578}.

\bibitem[Gretton et~al., 2012]{grettonKernelTwoSampleTest2012}
Gretton A., Borgwardt K.~M., Rasch M.~J., Schölkopf B., and Smola A. (2012).
\newblock A {Kernel} {Two}-{Sample} {Test}.
\newblock {\em Journal of Machine Learning Research}, 13(25):723--773.

\bibitem[Grootendorst et~al., 2021]{grootendorst2021maartengr}
Grootendorst M., Fuetterer H.-A., Luca F., Dhadse A., Matsak A., Pechersky I., Govil P., Frampton S., Ogura Y., et~al. (2021).
\newblock Maartengr/keybert: v0. 9.

\bibitem[Hamilton et~al., 2016]{hamiltonDiachronicWordEmbeddings2016}
Hamilton W.~L., Leskovec J., and Jurafsky D. (2016).
\newblock Diachronic {Word} {Embeddings} {Reveal} {Statistical} {Laws} of {Semantic} {Change}.
\newblock In Erk K. and Smith N.~A., editors, {\em Proceedings of the 54th {Annual} {Meeting} of the {Association} for {Computational} {Linguistics} ({Volume} 1: {Long} {Papers})}, pp. 1489--1501. Association for Computational Linguistics.

\bibitem[Han et~al., 2018]{hanConditionalWordEmbedding2018}
Han R., Gill M., Spirling A., and Cho K. (2018).
\newblock Conditional {Word} {Embedding} and {Hypothesis} {Testing} via {Bayes}-by-{Backprop}.
\newblock In Riloff E., Chiang D., Hockenmaier J., and Tsujii J., editors, {\em Proceedings of the 2018 {Conference} on {Empirical} {Methods} in {Natural} {Language} {Processing}}, pp. 4890--4895. Association for Computational Linguistics.

\bibitem[Jayasumana et~al., 2024]{jayasumana2024rethinking}
Jayasumana S., Ramalingam S., Veit A., Glasner D., Chakrabarti A., and Kumar S. (2024).
\newblock Rethinking fid: Towards a better evaluation metric for image generation.
\newblock In {\em Proceedings of the IEEE/CVF conference on computer vision and pattern recognition}, pp. 9307--9315.

\bibitem[Keidar et~al., 2022]{keidarSlangvolutionCausalAnalysis2022}
Keidar D., Opedal A., Jin Z., and Sachan M. (2022).
\newblock Slangvolution: A causal analysis of semantic change and frequency dynamics in slang.

\bibitem[Ledoit and Wolf, 2004]{ledoit2004well}
Ledoit O. and Wolf M. (2004).
\newblock A well-conditioned estimator for large-dimensional covariance matrices.
\newblock {\em Journal of Multivariate Analysis}, 88(2):365--411.

\bibitem[Liu et~al., 2021]{liu-etal-2021-statistically}
Liu Y., Medlar A., and Glowacka D. (2021).
\newblock Statistically significant detection of semantic shifts using contextual word embeddings.
\newblock In Gao Y., Eger S., Zhao W., Lertvittayakumjorn P., and Fomicheva M., editors, {\em Proceedings of the 2nd Workshop on Evaluation and Comparison of NLP Systems}, pp. 104--113, Punta Cana, Dominican Republic. Association for Computational Linguistics.

\bibitem[Mariani et~al., 2022]{marianiNLP4NLP+5DeepRevolution2022}
Mariani J., Francopoulo G., Paroubek P., and Vernier F. (2022).
\newblock {NLP4NLP}+5: {The} {Deep} ({R})evolution in {Speech} and {Language} {Processing}.
\newblock {\em Frontiers in Research Metrics and Analytics}, 7.

\bibitem[Periti and Montanelli, 2024]{peritiLexicalSemanticChange2024}
Periti F. and Montanelli S. (2024).
\newblock Lexical {Semantic} {Change} through {Large} {Language} {Models}: a {Survey}.
\newblock {\em ACM Computing Surveys}, 56(11):1--38.

\bibitem[Podolskiy et~al., 2021]{podolskiy2021revisiting}
Podolskiy A., Lipin D., Bout A., Artemova E., and Piontkovskaya I. (2021).
\newblock Revisiting mahalanobis distance for transformer-based out-of-domain detection.
\newblock In {\em Proceedings of the AAAI conference on artificial intelligence}, volume~35, pp. 13675--13682.

\bibitem[Robinson et~al., 2022]{robinson2022improvement}
Robinson B., Malinas R., Latimer V., Morrison B.~B., and Hero A.~O. (2022).
\newblock An improvement on the hotelling $t^2$ test using the ledoit-wolf nonlinear shrinkage estimator.
\newblock In {\em 2022 30th European Signal Processing Conference (EUSIPCO)}, pp. 2106--2110. IEEE.

\bibitem[Schlechtweg et~al., 2020]{schlechtwegSemEval2020Task2020}
Schlechtweg D., McGillivray B., Hengchen S., Dubossarsky H., and Tahmasebi N. (2020).
\newblock Semeval-2020 task 1: Unsupervised lexical semantic change detection.

\end{thebibliography}


\end{document}